%% file: template.tex
\documentclass[preprint,journal]{vgtc}  

\usepackage{amssymb}
\usepackage{xspace}
\PassOptionsToPackage{pagebackref,bookmarks}{hyperref}
\usepackage[colorinlistoftodos]{todonotes}
\usepackage{hyperref}
\usepackage{pdfcomment}
\usepackage{array}
\usepackage{makecell}
\usepackage{adjustbox}
\usepackage{xcolor}
\usepackage{tikz}
\newcommand{\data}{LegendBench\xspace}
\newcommand{\eg}{\textit{e.g.}\xspace}
\newcommand{\ie}{\textit{i.e.}\xspace}

\definecolor{XinnuoOrange}{HTML}{FF8C00}

\newcommand{\revise}[1]{\textcolor{black}{#1}}

\title{\data: A Diagnostic Benchmark for Legend Understanding with Counterfactual Interventions}

\author{%
Xinnuo Zhang,
Zhike Tang,
Jing Xu,
Haoyuan Zhao,
Weikai Yang\thanks{Corresponding author. E-mail: weikaiyang@hkust-gz.edu.cn}\\
The Hong Kong University of Science and Technology (Guangzhou)
}

\authorfooter{\item[]\mbox{}}

\preprinttext{arXiv preprint.}
\manuscriptnote{}

\input{0-Abstract}

\usepackage{svg}
\teaser{
  \centering

  \includegraphics[width=\linewidth, trim=33 50 29 20, clip]{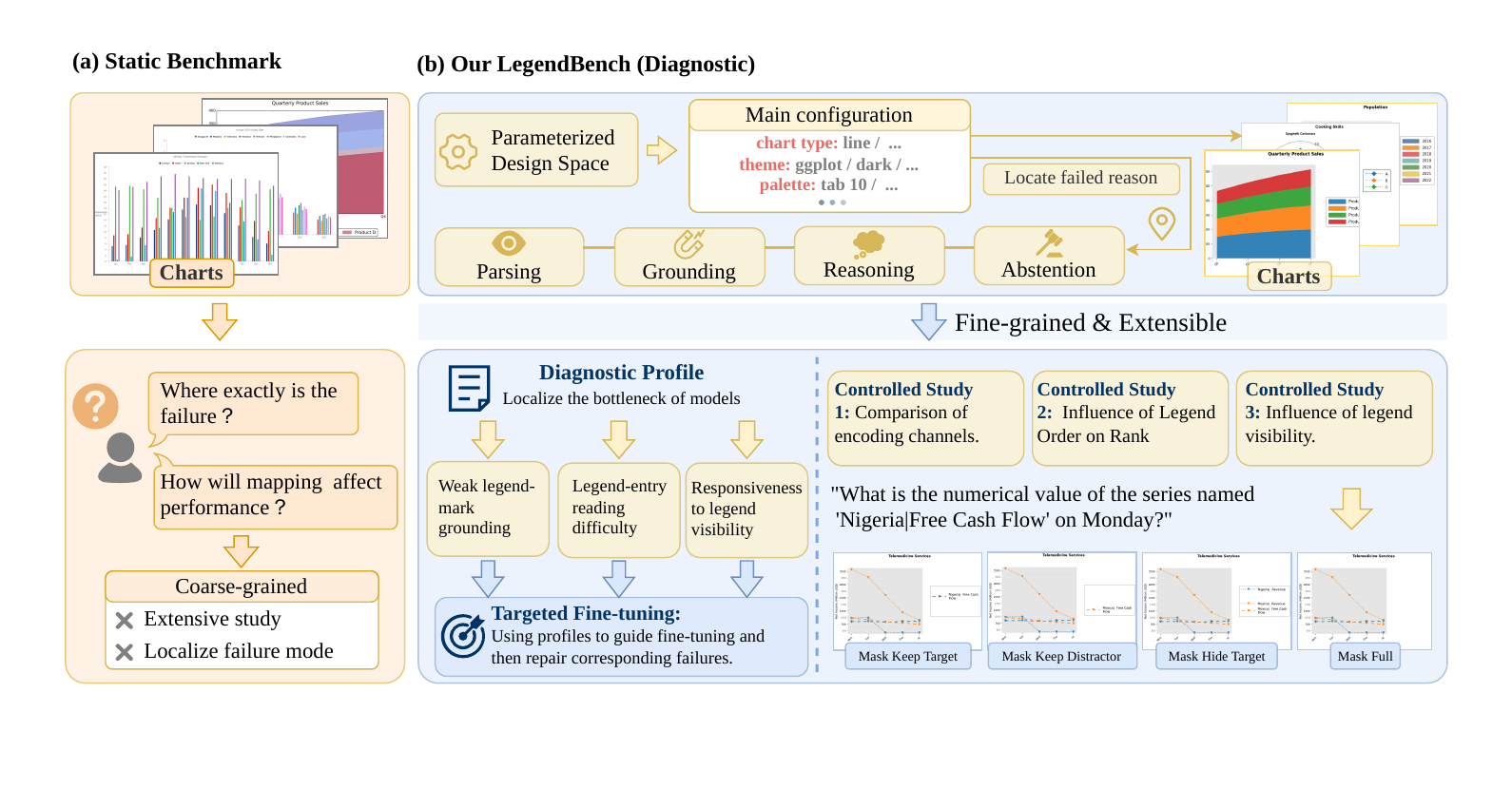}  
  \vspace{-25pt} 
  \caption{
    Overview of \data. Compared with static chart benchmarks, \data organizes evaluation around a parameterized design space, four legend-centric capability task families, and matched counterfactual variants. \revise{Their outcomes are summarized as a model-level capability profile,} enabling both failure localization and controlled legend interventions.
  }

  \label{fig:teaser}
}

\graphicspath{{figs/}{figures/}{pictures/}{images/}{./}} %

\usepackage{booktabs}                  %
\usepackage{lipsum}                    %
\usepackage{mwe}                       %
\usepackage{wrapfig}
\usepackage{mathptmx}                  %
\usepackage{listings}
\usepackage{amsmath}
\usepackage{tikz}
\usetikzlibrary{patterns}
\usepackage[table]{xcolor}
\usepackage{multirow}
\usepackage[utf8]{inputenc}
\usepackage{geometry}
\usepackage{nicematrix}
\usepackage{xfp}
\usepackage{colortbl} %
\usepackage{etoolbox}
\usepackage{pgf}
\usepackage{graphicx}
\usepackage{calc}
\DeclareMathAlphabet{\mathcal}{OMS}{cmsy}{m}{n}

\newcommand{\swatchH}{1.0ex}
\newcommand{\swatchW}{0.6em}

\newcommand{\markerS}{0.7em}
\newcommand{\lineW}{1.5em}
\newcommand{\palettebar}[1]{%
\raisebox{0.2ex}{%
\begingroup
\setlength{\fboxsep}{0pt}%

\endgroup
}}

\usepackage{tabularx}
\usepackage{array}
\PassOptionsToPackage{table,svgnames,dvipsnames}{xcolor}
\newcommand{\rowH}{\rule{0pt}{2pt}}
\renewcommand{\arraystretch}{1.1}
\definecolor{mr0}{HTML}{F0FAF6} %
\definecolor{mr4}{HTML}{55C4A0} %

\newcommand{\acc}[1]{%
    \cellcolor{mr4!\fpeval{(#1)^4*100}!mr0}%
    \makebox[2.4em][c]{\footnotesize #1}%
}

\newcolumntype{C}[1]{>{\centering\arraybackslash}m{#1}}
\newcolumntype{L}[1]{>{\raggedright\arraybackslash}m{#1}}
\newcommand{\colorbar}[1][Acc.]{%
    \begin{tikzpicture}[baseline=(current bounding box.center)]
        \shade[top color=mr4, bottom color=mr0] (0,0) rectangle (0.15, 1.8);
        \draw[lightgray] (0,0) rectangle (0.15, 1.8);
        \node[right, font=\tiny] at (0.15, 1.8) {1.0};
        \node[right, font=\tiny] at (0.15, 0.9) {0.5};
        \node[right, font=\tiny] at (0.15, 0) {0.0};
        \node[rotate=90, font=\tiny\bfseries] at (-0.3, 0.9) {#1};
    \end{tikzpicture}%
}

\definecolor{morandi1}{HTML}{E8D5F5}
\definecolor{morandi2}{HTML}{E8A97E}
\definecolor{morandi3}{HTML}{6BA3BE}
\definecolor{morandi4}{HTML}{82B89A}
\definecolor{morandi5}{HTML}{C9939A}

\definecolor{catpurple}{HTML}{A98BB5}
\definecolor{catteal}{HTML}{76B7B2}
\definecolor{catyellow}{HTML}{EDC948}
\definecolor{catgray}{HTML}{BAB0AC}

\definecolor{sblue1}{HTML}{6BA3BE}
\definecolor{sblue2}{HTML}{7FB7CF}
\definecolor{sblue3}{HTML}{93C4D9}
\definecolor{sblue4}{HTML}{A8D2E3}

\definecolor{sred1}{HTML}{C9939A}
\definecolor{sred2}{HTML}{D6A4AA}
\definecolor{sred3}{HTML}{E3B5BA}
\definecolor{sred4}{HTML}{F0C6CB}

\definecolor{tabblue}{HTML}{1F77B4}
\definecolor{taborange}{HTML}{FF7F0E}
\definecolor{tabgreen}{HTML}{2CA02C}
\definecolor{tabred}{HTML}{D62728}
\definecolor{tabpurple}{HTML}{9467BD}

\definecolor{setoneA}{HTML}{E41A1C}
\definecolor{setoneB}{HTML}{377EB8}
\definecolor{setoneC}{HTML}{4DAF4A}
\definecolor{setoneD}{HTML}{984EA3}
\definecolor{setoneE}{HTML}{FF7F00}

\definecolor{darktwoA}{HTML}{1B9E77}
\definecolor{darktwoB}{HTML}{D95F02}
\definecolor{darktwoC}{HTML}{7570B3}
\definecolor{darktwoD}{HTML}{E7298A}
\definecolor{darktwoE}{HTML}{66A61E}

\definecolor{settwoA}{HTML}{66C2A5}
\definecolor{settwoB}{HTML}{FC8D62}
\definecolor{settwoC}{HTML}{8DA0CB}
\definecolor{settwoD}{HTML}{E78AC3}
\definecolor{settwoE}{HTML}{A6D854}

\definecolor{rdbuA}{HTML}{B1182B}
\definecolor{rdbuB}{HTML}{F3A481}
\definecolor{rdbuC}{HTML}{F6F7F7}
\definecolor{rdbuD}{HTML}{90C4DD}
\definecolor{rdbuE}{HTML}{2065AB}

\definecolor{rdylgnA}{HTML}{D62F27}
\definecolor{rdylgnB}{HTML}{FDAD60}
\definecolor{rdylgnC}{HTML}{FFFFBF}
\definecolor{rdylgnD}{HTML}{A5D86A}
\definecolor{rdylgnE}{HTML}{199750}

\definecolor{simblueA}{HTML}{C1DBEF}
\definecolor{simblueB}{HTML}{7AB6D9}
\definecolor{simblueC}{HTML}{4191C6}
\definecolor{simblueD}{HTML}{1967AD}
\definecolor{simblueE}{HTML}{083C7D}

\definecolor{simredA}{HTML}{FFD1EC}
\definecolor{simredB}{HTML}{FB7656}
\definecolor{simredC}{HTML}{EE3A2C}
\definecolor{simredD}{HTML}{BF151B}
\definecolor{simredE}{HTML}{7E0610}

\begin{document}
\maketitle

\input{1-Intro}

\input{2-Related}

\input{3-LegendBench}

\input{3-2-LegendBenchConstruction}

\input{4-Experiments}

\input{4-2-Controlled}

\input{5-Discussion}

\input{6-Conclusion}

\section*{Supplemental Materials}
The supplementary document includes (1) additional benchmark examples and benchmark composition statistics, (2) further details of the generation pipeline, prompts, evaluation settings, and metric computation, and (3) complete encoding-channel, legend order shortcut and visibility-based abstention results for all tested models.
\bibliographystyle{abbrv-doi-hyperref}

\bibliography{main}

\end{document}

%% file: 0-Abstract.tex
\abstract{
Legends are fundamental to chart understanding, as reliable interpretation requires correctly binding legend entries to corresponding visual marks.
While vision-language models (VLMs) are increasingly applied to chart understanding, their legend understanding is poorly diagnosed by aggregate accuracy, which can be satisfied by superficial shortcuts and confound legend-specific errors with other reasoning failures.
To enable fine-grained diagnosis and controlled testing, we introduce \textit{\data}, a parametric benchmark and generation pipeline that produces targeted legend-centric test cases. %
\data contributes (1) a capability-task taxonomy spanning legend parsing, legend grounding, legend-conditioned reasoning, and legend-aware abstention to localize failures, and (2) counterfactual group generation, where each base chart yields multiple variants under controlled legend interventions to probe model invariance and sensitivity.
Using \data, we evaluate both general-purpose VLMs and specialized chart models \revise{and generate their capability profiles}, revealing persistent bottlenecks in reliable legend-to-mark binding and counterfactual consistency.
\revise{We then use these capability profiles to guide targeted fine-tuning, demonstrating that bottleneck-specific interventions can effectively close the localized capability gaps and generalize to unseen data. %
We further leverage our counterfactual design to conduct fine-grained diagnostic experiments, analyzing encoding-channel effects, legend-order shortcuts, and abstention under varying visibility.}
}

%% file: 1-Intro.tex
\section{Introduction}
\label{sec:intro}

Data visualizations communicate quantitative information by encoding data in visual variables such as color, marker shape, and line style.
Among chart components, the \emph{legend} is essential because it declares how those encodings map onto \revise{named series or quantitative scales.}
\revise{A mark becomes interpretable only once it is bound to that mapping, so legend understanding is a prerequisite for reliable chart reading.}
Although vision-language models (VLMs) can now answer many questions over charts~\cite{chartqa,encqa,chartmuseum}, a correct answer does not mean it is a reliable one.
\revise{It may instead exploit shortcuts that skip legend-mark binding, such as treating the position of a legend entry as a proxy for the visual rank of the corresponding series.
Aggregate accuracy cannot tell these cases apart, because a shortcut can produce the same answer as genuine legend understanding until the legend is changed.}
We therefore argue that evaluating chart understanding requires testing whether a model's answers track the legend directly, for instance whether reordering or swapping legend entries changes its answers as it should, rather than measuring correctness alone.

\revise{Legends provide a particularly useful probe for such evaluation because their mappings can be isolated and systematically manipulated.
In this paper, we use \emph{legend} in a broad but explicit sense: visual mechanisms that specify how semantic referents map to encodings.
This includes conventional categorical legend boxes, direct labels attached to marks, and continuous scale guides such as color bars on heatmaps.
We focus on legend-mediated understanding for three reasons.
First, such legend-mediated mappings are necessary for reading multi-series charts, and an incorrect mapping can therefore propagate to all subsequent reasoning over the chart rather than causing a single localized error.
Second, the mapping is spatially separated from the marks it explains, so using it correctly requires binding across regions of the image.
Third, the legend can be edited without touching the underlying data, so a fully parameterizable, matched intervention can isolate whether a model genuinely relies on legend information.
Together, these properties make legends a high-leverage diagnostic probe for testing robust chart understanding.}

To this end, we introduce \textbf{\data}, a \revise{legend-centric diagnostic} for assessing whether model behavior is grounded in the declared legend mapping.
\revise{A reliable reader should parse the declared mapping, bind it to the corresponding marks or scale values, remain consistent under mapping-preserving interventions, update its answer under mapping-altering interventions, and abstain when the required evidence is absent or ambiguous.
Informed by cognitive models of graph comprehension~\cite{pinker1990,carpentershah1998}, \data organizes these capabilities into four task families: legend parsing, legend grounding, legend-conditioned reasoning, and legend-aware abstention, each further decomposed into fine-grained sub-capabilities.
Rather than providing a static collection of examples, \data is supported by a parametric generation pipeline that factorizes chart specification, legend specification, and question--answer pairs.
This factorization enables controlled generation across chart types, legend designs, and question types, as well as matched counterfactual variants produced through selective legend interventions while other chart factors are held fixed.
These variants test complementary behavioral requirements, \ie, predictions should remain invariant under mapping-preserving transformations and change appropriately under mapping-altering interventions.
Comparing behavior across matched variants therefore exposes failures that aggregate accuracy cannot distinguish, including differential sensitivity to encoding channels, shortcut reliance, and inappropriate responses when legend evidence is insufficient.}

We evaluate current general-purpose and chart-specialized VLMs as a worked example of this diagnostic workflow.
\revise{The resulting capability profiles reveal a recurring gap between reading a legend and using it correctly, and models are considerably more consistent on mapping-preserving interventions than responsive to mapping-altering ones. %
We then use these profiles to guide targeted fine-tuning, demonstrating that training on diagnosed bottlenecks repairs the localized failures more effectively than other strategies, and the gains transfer beyond \data itself.
Moreover, we leverage our parametric, counterfactual design to conduct several fine-grained diagnostic experiments, analyzing encoding-channel effects, legend-order shortcuts, and abstention under varying visibility.}
Together, these results show that models with similar aggregate accuracy can fail for qualitatively different reasons, and that identifying these reasons can provide actionable guidance for model evaluation and improvement.
Our contributions are:
\begin{itemize}[nosep]
    \item A \revise{legend-centric capability taxonomy that decomposes legend-mediated understanding} into legend parsing, legend grounding, legend-conditioned reasoning, and legend-aware abstention.
    \item A parametrized generation pipeline that renders counterfactual chart variants under controlled legend interventions.
    \item \revise{A demonstration that \data drives diagnosis-guided improvement, revealing \emph{where} and \emph{why} state-of-the-art VLMs fail and guiding model selection and bottleneck-targeted fine-tuning in ways aggregate accuracy cannot.}
\end{itemize}

%% file: 2-Related.tex
\section{Related Work}
\label{sec:related_work}

\subsection{Visualization Literacy and Legend Comprehension}

Human-centered studies treat chart understanding as a structured competence through foundational literacy assessments \cite{boy2014,vlat,minivlat}, expanded measurement instruments \cite{calvi,adaptivelit,avec,mdamv}, analyses of comprehension barriers and higher-level interpretation \cite{barrierslit,highlevelcomp}, and legend-specific design studies \cite{interactivelegends,arealegends}.
Boy et al.~\cite{boy2014}, VLAT~\cite{vlat}, and Mini-VLAT~\cite{minivlat} frame chart reading as a measurable competence rather than a by-product of generic intelligence or numeracy.
CALVI~\cite{calvi}, Adaptive Assessment~\cite{adaptivelit}, AVEC~\cite{avec}, and MdamV~\cite{mdamv} broaden this view by covering misleading charts, adaptive testing, visual encoding construction, and multidimensional understanding.
Studies on literacy barriers~\cite{barrierslit} and high-level comprehension~\cite{highlevelcomp} show that chart-reading errors arise from multiple sources, including encoding misreadings, label--value confusion, and difficulty extracting chart messages.
Interactive legends~\cite{interactivelegends} and area-to-value legend studies~\cite{arealegends} further show that legends affect attention, lookup cost, and decoding accuracy rather than acting as peripheral annotations.
Together, these human-centered studies motivate treating chart understanding as a collection of separable subskills and legend comprehension as a distinct source of difficulty.
\revise{Stokes and Hearst~\cite{Stokes2026} provide the closest recent VIS taxonomy of labeling, analyzing text functions that identify mappings, while our taxonomy isolates the legend-mediated subset of that problem and makes it operational for counterfactual evaluation.}

Model-centered work increasingly transfers this literacy perspective to the evaluation of vision--language models through visualization-literacy tests, standardized human-centered instruments, and interpretability-oriented analyses \cite{gpt4vislit,standardizedvislit,chart6,probingvislit}.
The GPT-4 evaluation~\cite{gpt4vislit} shows that a strong general-purpose multimodal model remains vulnerable to precise reading, color-based judgments, and misleading designs.
Benchmarking Visual Language Models on Standardized Visualization Literacy Tests~\cite{standardizedvislit} brings standardized human literacy instruments directly into VLM evaluation.
CHART-6~\cite{chart6} shows that model errors differ systematically from human errors across multiple visualization-understanding tasks.
Probing the Visualization Literacy of Vision-Language Models~\cite{probingvislit} complements answer-level evaluation with analyses of what evidence models attend to.
Taken together, these model-centered studies strengthen the case for evaluating chart understanding at the level of subskills and evidence use rather than through a single aggregate score.

\subsection{Chart Understanding Benchmarks}

Chart understanding benchmarks have evolved from early controlled chart QA datasets \cite{figureqa,dvqa} to real-world, open-ended, and low-level task suites \cite{chartqa,opencqa,chartinsights,charttotext,vistext}, and then to broader collections that increase coverage, realism, and reasoning complexity \cite{chartbench,charxiv,chartmind,multichartqa}.
FigureQA~\cite{figureqa} and DVQA~\cite{dvqa} established chart question answering in controlled synthetic settings, which LEAF-QA~\cite{leafqa} and PlotQA~\cite{plotqa} later extended to richer chart types and stronger numerical reasoning.
This synthetic line was followed by ChartQA~\cite{chartqa}, which brought human-written questions to real-world charts, and by OpenCQA~\cite{opencqa}, Chart-to-Text~\cite{charttotext}, and VisText~\cite{vistext}, which expanded evaluation from short answers to open-ended responses and chart descriptions.
ChartInsights~\cite{chartinsights} shows that even low-level chart-reading operations remain difficult for current multimodal models, suggesting that broader benchmarks can still mask foundational weaknesses.

Subsequent benchmark design emphasized scale and realism.
ChartBench~\cite{chartbench} increases chart coverage through programmatic construction, while CharXiv~\cite{charxiv}, ChartMind~\cite{chartmind}, and MultiChartQA~\cite{multichartqa} move toward in-the-wild scientific charts, broader reasoning demands, and multi-chart scenarios.
Across this progression, benchmark design has steadily improved scope and difficulty, but the dominant evaluation signal is still end-task correctness, leaving open which visual subskills or evidence sources actually support correct answers.

\subsection{Diagnostic Evaluation Beyond Aggregated Accuracy}

Visual analytics has long sought to expose model internals and localize failure modes to support model refinement \cite{liu2017towards,yang2024foundation}. In chart understanding, a more recent line of work pursues a related diagnostic goal through chart-specific diagnostic benchmarks \cite{encqa,chartmuseum,fugu}, evidence-alignment and grounding analyses \cite{chartlens,chartab,splitground,groundingvlm}, and controlled intervention frameworks \cite{cfvlm,causalvlbench,treblecfvlm}.
EncQA~\cite{encqa} reorganizes chart evaluation around visual encodings and analytic tasks, showing that performance varies sharply across encoding--task pairs and that many benchmark items can be solved from text alone.
ChartMuseum~\cite{chartmuseum} extends this concern to real-world charts by separating questions according to visual versus textual reliance and reporting large drops once textual shortcuts are reduced.
FUGU~\cite{fugu} pushes diagnosis further down the stack by isolating foundational spatial skills and tracing failures to bottlenecks in spatial extraction and the vision--language handoff.
These works show that accuracy alone cannot reveal whether a model is using the intended visual evidence.

Once such failures are exposed at the task level, the next question is whether models align their answers with the right evidence.
ChartLens~\cite{chartlens} studies chart answers through fine-grained visual attribution, and ChartAB~\cite{chartab} evaluates dense grounding and alignment over chart elements.
SplitGround~\cite{splitground} and broader studies of visual grounding in VLMs~\cite{groundingvlm} argue that answer correctness should therefore be paired with explicit evidence alignment rather than inferred from final outputs alone.
However, attribution and grounding analyses remain largely observational.
CF-VLM~\cite{cfvlm}, CausalVLBench~\cite{causalvlbench}, and Treble~\cite{treblecfvlm} show that controlled interventions and causal probes provide a stronger test by checking whether predictions track semantically relevant changes instead of superficial correlations.
Together, these studies move evaluation beyond aggregate correctness toward finer-grained diagnosis, but they still leave open a chart-native framework that isolates distinct subskills and verifies, under controlled interventions, whether model predictions follow the intended visual evidence rather than textual or positional shortcuts.

%% file: 3-LegendBench.tex
\section{\data Design}
\data is designed as a diagnostic benchmark.
Its goal is not merely to count correct answers of the model, but to test whether the model \revise{truly utilizes the legend information and further identify at which step the legend understanding process fails.}

\begin{table*}[t]
\centering
\footnotesize
\setlength{\tabcolsep}{4pt}
\renewcommand{\arraystretch}{1.0}
\setlength{\extrarowheight}{5pt}
\caption{Parameterized space of a \data base unit $b=(s,l,q,y)$.
Separating chart-side, legend-side, and question-side factors enables matched regeneration for diagnostic benchmarking and counterfactual intervention.}
\begin{tabular}{>{\columncolor{morandi2!30}}m{3.3cm}
                >{\columncolor{morandi4!30}}m{2.5cm}
                >{\columncolor{morandi5!30}}m{5.8cm}
                >{\columncolor{morandi3!30}}m{5.3cm}}
\toprule
\rowcolor{morandi1!42}
\textbf{Group} & \textbf{Variable} & \textbf{Examples} & \textbf{Role} \\
\midrule
Chart Specification ($s$) & chart families
& line, scatter, radar, area, pie, bar, heatmap
& defines the chart family and compatible mark families \\

& semantic labels
& entity categories, metric categories
& defines the semantic meaning of each series, especially in compositional encoding mode \\

& structural complexity
& num\_series, num\_points
& controls legend cardinality and chart density \\

& visual presentation& theme\_name, AddLabels & changes non-legend appearance and chart-side text support while preserving chart semantics \\

\midrule
Legend Specification ($l$)& encoding channel& color, marker, line style, texture& \revise{distinguishes series identity or, for a color bar, the quantitative scale}\\

 & encoding mode& single, double, compositional& defines how channels encode series identity\\

& \revise{legend form}& \revise{categorical box, direct labels, continuous color bar}& \revise{specifies how the mapping is exposed}\\

& legend style& color\_scheme, marker\_scheme, linestyle\_scheme, texture\_scheme, legend\_position, legend\_size, layout\_cols, visibility& modifies $l$ while keeping $s,q$ fixed whenever possible\\

\midrule
Question-Answer Pair $(q,y)$
& task family
& parsing, grounding, reasoning, abstention
& determines what capability is evaluated \\

& question subfamily
& legend reading / semantic-visual mapping / channel recognition; value extraction; computation / graph reasoning; masked answerability
& controls benchmark coverage and diagnostic focus \\

\bottomrule
\end{tabular}

\label{tab:legendbench_designspace}
\end{table*}

\subsection{Design Requirements}

\revise{Based on the evaluation gaps identified in Sec.~\ref{sec:related_work}, we identify the following design requirements that the benchmark must satisfy.}

\textbf{R1. Support Failure Localization.}
\revise{Reliable legend understanding requires localizing where the process breaks, because a correct final answer can still hide a failed earlier stage.
Pinker~\cite{pinker1990} and Carpenter and Shah~\cite{carpentershah1998} describe chart reading as a staged process of encoding, binding, and inference.
Legend use inherits this structure:} a model may fail to read the declared mapping, bind it to the wrong marks, err in downstream reasoning after correct grounding, or guess when the necessary legend information is unavailable.
These errors should not be collapsed into one accuracy number.
Instead, a useful benchmark must decompose them into individually testable targets. 

\textbf{R2. Being Parameterized and Extensible.}
Each of the stages above is affected by multiple interacting variables, such as chart types, series count, and rendering style.
These variables are typically confounded in a static chart collection.
To attribute model behavior to specific design choices, the benchmark must factorize these variables into a parameterized specification so that each can be varied independently while the others are held fixed.
This also makes the benchmark extensible, as new chart types, intervention types, or question templates can be incorporated without redesigning the pipeline.

\textbf{R3. Support Causal Testing.}
Even with fine-grained tasks and a parameterized space, a correct answer on a single chart--question pair does not guarantee that the model correctly used the legend.
Instead, it may result from shortcuts such as color or position biases.
To distinguish genuine legend grounding from such shortcuts, the benchmark 
must support controlled causal analysis by producing matched counterfactual variants that manipulate only the legend while holding the underlying data and question intent fixed.

These three requirements map directly to three key designs of \data.
R1 motivates a legend-centric task taxonomy that decomposes legend understanding into observable targets.
R2 motivates a parameterized chart--legend space in which the relevant variables are explicit and independently controllable.
R3 motivates counterfactual evaluation groups that test whether legend information is used causally.

\subsection{Legend-Centric Task Taxonomy}
\label{sec:taxonomy}

To satisfy R1, the benchmark must separate the major stages of legend understanding into individually testable targets, so outcomes can be interpreted as \emph{where} the process breaks rather than only \emph{whether} the final answer is correct.
\revise{Those stages are named in R1: encoding, binding, and inference~\cite{pinker1990,carpentershah1998}, together with judging when the required legend evidence is unavailable.
We thus operationalize them as four legend-centric task families rather than as an independent question list.}

\textbf{T1: Legend Parsing.}
This task evaluates whether a model can read the mapping that the legend declares.
\revise{For a categorical box, this means recovering each label--visual pairing inside the legend region.
For direct labels, the model reads the local label text adjacent to a chart mark.
For a continuous color bar, this means reading scale endpoints, ticks, and orientation.}
This is the foundation for all subsequent tasks.

\textbf{T2: Legend Grounding.}
While T1 operates within the legend region, T2 evaluates whether the model can use that mapping in the plot.
\revise{For categorical legends and direct labels, the model must locate the corresponding marks and bind them to the correct entries.
For a continuous color bar, the model must map a cell or mark color through the bar to a quantitative value.
Value extraction is a separate downstream grounding sub-capability because it depends both on correct target binding and on quantitative scale reading.}

\textbf{T3: Legend-Conditioned Reasoning.}
This task evaluates downstream chart understanding when the answer depends on correct legend grounding, such as comparing series, detecting intersections, or performing multi-step computations.
A model can only solve T3 reliably if it has already succeeded on T2.

\textbf{T4: Legend-Aware Abstention.}
This task evaluates whether a model can recognize that a question is unanswerable when the required legend information is missing, occluded, or conflicting, rather than producing a confident but unfounded guess.

\revise{T1--T3 follow that encoding--binding--inference order, so a matched diagnostic chain can explicitly name the earliest failed stage rather than only the final error.
T4 serves as a cross-cutting evidence-sufficiency check on whether the model withholds an answer when the evidence required by that chain is missing.}

\subsection{Parameterized Space}
\label{sec:attributes}

To satisfy R2, we factorize each benchmark instance into a set of explicit, independently controllable dimensions.
We define a base unit as $b=(s, l, q, y)$, where:

\begin{itemize}[nosep]

\item \textbf{Chart Specification ($s$)} contains the underlying data and its visual realization. Legend-related information is excluded. 

\item \textbf{Legend Specification ($l$)} is the core dimension of \data.
It specifies \revise{the visual mapping represented by the legend and how this mapping is presented, such as through a categorical box, direct labels, or a continuous color bar.}

\item \textbf{Question and Answer Pair($q,y$)} specifies the evaluation question and its expected answer under intervention.
\end{itemize}

Taken together, this base unit $b=(s, l, q, y)$ defines a \emph{parameterized space} rather than a static dataset template.
This space serves two purposes:
1) it specifies which variables can be systematically varied to study legend understanding in isolation from unrelated factors, and 
2) it provides the foundation for controlled analyses beyond simple legend intervention.
Table~\ref{tab:legendbench_designspace} summarizes the concrete controllable variables in our current instantiation of this design space.

\begin{figure*}[t]
\centering
\includegraphics[width=\textwidth,
trim=23 200 20 20, clip]{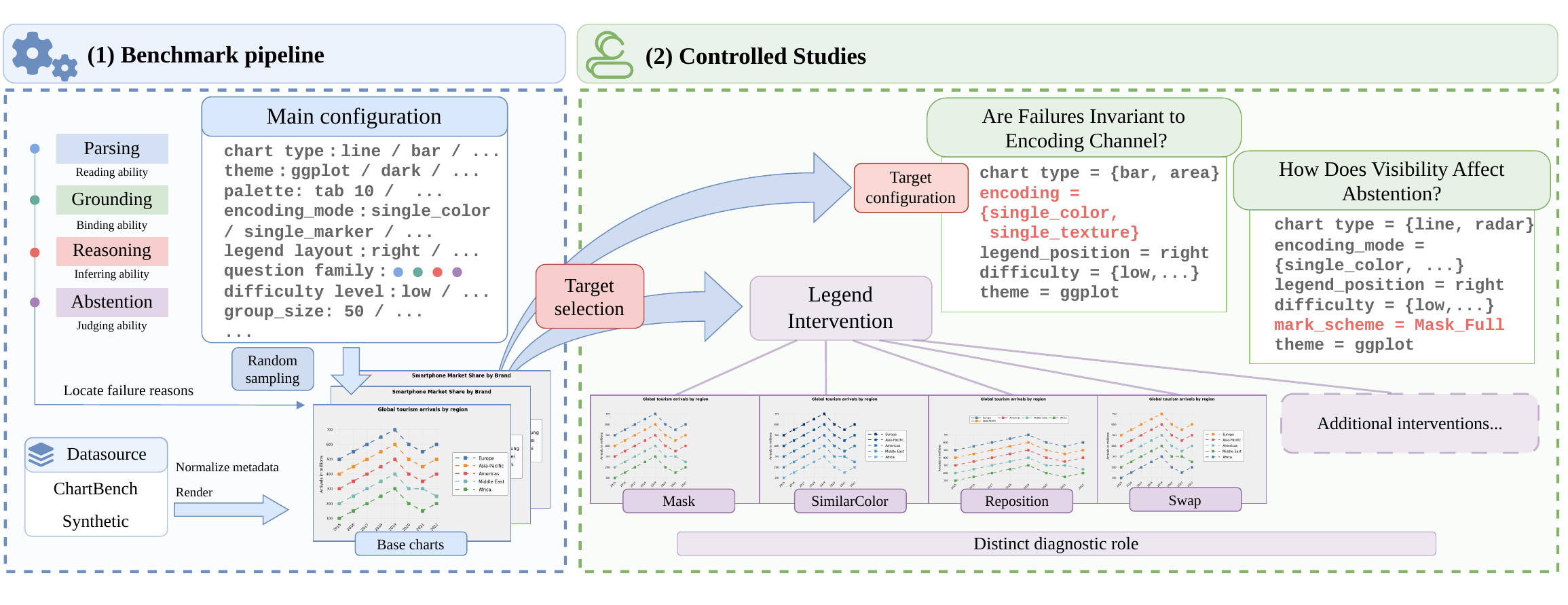}
\caption{Instantiation pipeline of \data. A main configuration defines each base chart instance, while target configurations specify the dimensions selected for controlled diagnostic experiments. The pipeline then applies legend interventions to generate matched variants for diagnostic evaluation.}
\label{fig:pipeline}
\end{figure*}

\subsection{Counterfactual Evaluation Group}
\label{sec:counterfactual}

The task taxonomy and parameterized space above tell us \emph{what} to measure and \emph{which variables} to control, but evaluating each base unit $b$ in isolation cannot reveal whether a correct answer genuinely depends on the legend (R3).
We therefore organize evaluation around \emph{counterfactual groups}: given a base unit $b=(s,l,q,y)$, we construct a group $\mathcal{G}(b)$ by applying intervention operators primarily to the legend specification $l$, producing variants $(s,l',q',y')$ that hold the chart-side fixed.
Any change in model behavior across the group can therefore be attributed to the legend manipulation rather than to incidental differences in chart content.
Note that the question $q$ and ground-truth $y$ may vary due to the controlled intervention on $l$.
For example, if the question is ``What is the highest value in the red line?'', changing the legend--mark binding from A$\leftrightarrow$red to A$\leftrightarrow$blue would change the question to ``What is the highest value in the blue line?'' so that the difficulty of the question remains the same.

In this sense, the counterfactual group directly enables us to test whether model predictions are causally dependent on controlled legend information rather than supported by shortcut cues.
At the same time, it is a reusable intervention abstraction that can support not only the current legend manipulations, but also future extensions over other controlled dimensions in the parameterized space.

%% file: 3-2-LegendBenchConstruction.tex
\section{\data Construction}
\label{sec:dataset}
Following the parameterized space of \data, we now describe the \textit{instantiation pipeline} that constructs each base evaluation unit $b = (s,l,q,y)$.
\revise{It is the factorization of chart specification, legend specification, and question--answer pair, together with matched counterfactual groups that update $q$ and $y$ when only the legend changes.
By representing each dimension as a discrete, controllable parameter, every instance remains a structured object whose visual realization can be modified while its semantic intent stays synchronized (Figure~\ref{fig:pipeline}).}

\subsection{Chart Specification Generation ($s$)}
The instantiation process begins with the chart specification $s$, which serves as the foundational skeleton of a benchmark instance while explicitly excluding any legend specification.
This separation is essential to our benchmark design: by keeping the chart scaffold independent from the legend--mark binding, we can later apply controlled  legend interventions to individual chart instances without changing the underlying data values or the intended analytical task.

\begin{table*}[t]
\centering
\footnotesize
\setlength{\tabcolsep}{4pt}
\renewcommand{\arraystretch}{1.0}
\setlength{\extrarowheight}{5pt}
\caption{\revise{Capability tasks across legend forms, each of which exposes different parsing evidence and requires
different grounding operations.}
}
\begin{tabular}{
    >{\columncolor{morandi2!30}\raggedright\arraybackslash}m{3.0cm}
    >{\columncolor{morandi4!30}\raggedright\arraybackslash}m{3.9cm}
    >{\columncolor{morandi5!30}\raggedright\arraybackslash}m{4.5cm}
    >{\columncolor{morandi3!30}\raggedright\arraybackslash}m{5.6cm}
}
\toprule
\rowcolor{morandi1!42}
\textbf{Legend form}
& \textbf{Parsing evidence}
& \textbf{Grounding operation}
& \textbf{Typical chain} \\
\midrule

Categorical box
& Labels and visual tokens presented in the legend box
& Bind a legend entry to the corresponding plotted mark or series
& \mbox{Parsing-Binding-Reasoning} or \mbox{Parsing-Binding-Value Extraction-Reasoning} \\

Direct labels
& Local label text placed next to a plotted mark
& Associate the label with the adjacent plotted mark
& \mbox{Parsing-Binding-Reasoning} or \mbox{Parsing-Binding-Value Extraction-Reasoning} \\

Continuous color bar
& Scale endpoints, ticks, orientation, and quantitative range
& Localize the target and map its color through the continuous scale
& \mbox{Parsing-Localization-Value Extraction-Reasoning} \\

\bottomrule
\end{tabular}
\label{tab:guideform_capability_tasks}
\end{table*}

\textbf{Chart Types.}
To cover common legend-mediated visualizations, we organize the implementation by mark family: (1) point-based (\texttt{Scatter}), (2) line-based (\texttt{Line}, \texttt{Radar}), and (3) area-based (\texttt{Bar}, \texttt{Pie}, \texttt{Area}, \revise{\texttt{Heatmap}}).
This organization exposes different grounding demands.
Point-based charts emphasize discrete marker identity, line-based charts emphasize patterns along trajectories, and area-based charts emphasize filled regions.
By organizing charts around mark families rather than isolated chart names, \data can test whether legend comprehension is specific to one visual grammar.

\textbf{Underlying Data.}
Once the chart type is determined, we instantiate the underlying data through a hybrid sourcing strategy.
First, we adapt structured data tables from ChartBench~\cite{chartbench}.
Second, we complement these external samples with a synthetic generator in order to guarantee fine-grained parametric control over semantic composition, structural complexity, and visual difficulty.
Across synthetic samples, semantic series labels are generated in an \textit{entity}~$\mid$~\textit{metric} format (\eg \textit{NVIDIA $\mid$ Net Profit Margin}) so that legend entries remain semantically plausible while preserving explicit compositional structure.
For line- and area-based charts, we use structured trajectory templates such as linear, exponential, S-curve, and event-driven patterns so that questions have known structure.
For other chart families, the generator uses chart-specific value instantiation that matches the visual grammar.

\textbf{Structural Complexity.}
In addition to value generation, the data stage also determines key structural factors such as the number of series and the number of points per series.
Hence, we explicitly parameterize the structural complexity within $s$. By controlling variables such as \texttt{num\_series} and \texttt{num\_points}, we can systematically stratify samples by difficulty.
These parameters directly influence chart density and legend cardinality, and therefore contribute to the overall difficulty of the instance.
By fixing them before the legend specification, the chart side remains fixed in subsequent interventions, ensuring that any observed behavioral changes can be attributed to the legend operation.

\textbf{Visual Presentation.}
To guarantee visual diversity and to evaluate whether models rely on explicit textual grounding, we explicitly parameterize chart appearance through the \texttt{theme\_name} attribute and the presence or absence of data-level annotations through an \texttt{AddLabels} setting.
The \texttt{theme\_name} parameter controls global style properties such as background tone, grid visibility, and axis appearance. These changes do not alter the underlying data or legend semantics, but they do change the broader visual context.
The \texttt{AddLabels} setting controls whether the corresponding values are directly written onto the chart marks. When the labels are visible, some problems may become easier because the model can rely more on local textual cues, reducing the global search burden of the model.
By changing these representation-level attributes independently of the underlying chart and legend structure, benchmarking can test whether the performance reflects that the model has a robust legend understanding or relies on auxiliary text support.

\subsection{Legend Specification Generation ($l$)}  
With the chart specification ($s$) fixed, the pipeline instantiates the legend specification $l$, which dictates how series identity \revise{or a quantitative scale} is exposed.
Most existing chart benchmarks treat the legend as pixels in a rendered image, which makes it difficult to know what changed when the legend is edited.
Our pipeline represents the legend as an independent specification\revise{, including three legend forms: a categorical box, direct labels, and a continuous color bar.}
This representation lets the generator apply the same intervention operators to content, presentation, and visibility.

\textbf{Encoding Channels}.
We assign visual tokens to each series using four fundamental channels: \textit{color}
  \raisebox{-0.1ex}{\tikz\draw[morandi1,fill=morandi1,line width=0.5pt](0,0)rectangle(\swatchW,\swatchH);}%
  \raisebox{-0.1ex}{\tikz\draw[morandi2,fill=morandi2,line width=0.5pt](0,0)rectangle(\swatchW,\swatchH);}%
  \raisebox{-0.1ex}{\tikz\draw[morandi3,fill=morandi3,line width=0.5pt](0,0)rectangle(\swatchW,\swatchH);},
\textit{marker}
  \raisebox{-0.15ex}{\tikz\draw[morandi4,fill=morandi4!55,line width=0.7pt](0,0)circle(0.38em);}%
  \hspace{0.03em}
  \raisebox{-0.15ex}{\tikz\draw[morandi5,fill=morandi5!55,line width=0.7pt](0,0)rectangle(\markerS,\markerS);}%
  \hspace{0.03em}
  \raisebox{-0.15ex}{\tikz\draw[morandi2,fill=morandi2!55,line width=0.7pt](0.35em,0em)--(0.7em,0.35em)--(0.35em,0.7em)--(0em,0.35em)--cycle;},
\textit{line style}
  \raisebox{0.35ex}{\tikz\draw[morandi3,line width=1.4pt,line cap=round](0,0)--(\lineW,0);}%
  \hspace{0.05em}
  \raisebox{0.35ex}{\tikz\draw[morandi2,line width=1.4pt,dashed,line cap=round](0,0)--(\lineW,0);},
and \textit{texture}
  \raisebox{-0.1ex}{\tikz{
    \fill[morandi2!15](0,0)rectangle(\swatchW,\swatchH);
    \fill[pattern=north east lines,pattern color=morandi2](0,0)rectangle(\swatchW,\swatchH);
    \draw[morandi2,line width=0.5pt](0,0)rectangle(\swatchW,\swatchH);}}%
  \raisebox{-0.1ex}{\tikz{
    \fill[morandi3!15](0,0)rectangle(\swatchW,\swatchH);
    \fill[pattern=dots,pattern color=morandi3](0,0)rectangle(\swatchW,\swatchH);
    \draw[morandi3,line width=0.5pt](0,0)rectangle(\swatchW,\swatchH);}}%
  \raisebox{-0.1ex}{\tikz{
    \fill[morandi4!15](0,0)rectangle(\swatchW,\swatchH);
    \fill[pattern=horizontal lines,pattern color=morandi4](0,0)rectangle(\swatchW,\swatchH);
    \draw[morandi4,line width=0.5pt](0,0)rectangle(\swatchW,\swatchH);}}.
These channels were chosen because they cover common ways to distinguish categorical series and differ in perceptual difficulty~\cite{bertin1983,clevelandmcgill1984}.
Applicability depends on the mark family (\eg textures are applied to area-based charts, while markers are reserved for point-based or line-based charts). 
To achieve coverage of visual patterns, the pipeline samples predefined channel schemes (\texttt{color\_scheme}, \texttt{linestyle\_scheme}, \texttt{marker\_scheme}, \texttt{texture\_scheme}).

\textbf{Encoding Mode}:
To test whether models can process redundant or multi-dimensional cues, we support three encoding modes:
\begin{itemize}[nosep]
    \item \textbf{Single}: one visual channel encodes series identity;
    \item \textbf{Double}: two channels encode the same series identity;
    \item \textbf{Compositional}: two channels encode different semantic dimensions, \eg, color for \textit{Region} and marker for \textit{Product}.
\end{itemize}

The channel space and encoding mode space allow the coverage of benchmark tests to range from simple single-thread charts to more complex multi-channel graphs,
 which reflects the diversity of legend design and enables us to distinguish between errors caused by specific channels and those caused by the interaction between channels.

\textbf{Legend Style}:
In addition to semantic mapping, legend understanding also depends on whether the legend is easy to locate, easy to read, and easy to compare with the chart marks. We therefore explicitly parameterize these presentation-level variables (\eg \texttt{legend\_position}, \texttt{legend\_size}, \texttt{layout\_cols}, and \texttt{visibility}). Their role is to change the perceptual accessibility of the legend without altering the underlying data mapping itself. In the experiments, these controls are instantiated through variants such as \texttt{Reposition}, \texttt{ScaleDown}, and the visibility settings underlying \texttt{Mask-A} and \texttt{Mask-U}. This is important because such variations are common in real charts: legends may appear on the right, above the plot, in more compact layouts, or under reduced size and partial occlusion. By including these factors, the benchmark can test whether models remain stable under realistic changes in legend presentation rather than only under ideal layouts.

\subsection{Question-Answer Pair Generation ($q,y$)}
Once the chart-specification and the legend-specification are fixed, we generate question-answer pairs.
This stage defines what the model is required to do and what the correct response is.
In \data, the questions and answers are not fixed but are bound to the structured metadata of the instance.
This ensures that $q$ and $y$ stay aligned after an intervention.
We design the pairs to localize the stage at which legend understanding fails, rather than to score generic chart reading.

\textbf{Template generation.}
The question-generation system uses a template-based routing mechanism that sends each chart to a chart-specific generator.
This modular design lets questions match the visual structure of different chart families.
Question templates are not stored as static text.
Instead, they are instantiated from instance metadata, such as the target series label or the relevant data values.
Each question also includes a visual vocabulary hint that lists the visual channels used in the chart.
This reduces perceptual ambiguity without revealing the actual legend-mark mapping.
For instance, a Semantic-to-Visual question asks \textit{What visual properties are assigned to `Grade 3'?}, requiring the model to look up the legend and return the full visual specification.
A grounding question such as \textit{``What is the value of `Finance' in 2015?''} is a value extraction readout: it fails if the model binds the wrong mark, even though the surface form is a simple lookup.
\revise{Binding and value extraction are therefore recorded as separate grounding sub-capabilities, including continuous color-bar reading on heatmaps.}

\revise{\textbf{Diagnostic chain generation.}
For each chart, semantic target, and final question intent, the generator assigns matched capability-task questions from the generated questions. 
Each chain contains only the upstream probes required by its final question. 
As summarized in \Cref{tab:guideform_capability_tasks}, the required upstream operations depend on the legend form.
Relational questions typically use Parsing--Binding--Reasoning chains, whereas numerical questions use Parsing--Binding--Value Extraction--Reasoning chains. 
For continuous guides, target localization replaces discrete identity binding. Each question is answered independently, and model answers are not passed from one question to the next. 
A chain is counted as correct only when all applicable questions in the matched set are answered correctly.}

\textbf{Intent Preserving.}
A core strength of our pipeline is the ability to maintain semantic alignment across counterfactual groups. Each template is equipped with a vocabulary replacement mechanism: suppose a question points to a series through its visual appearance. If the legend mapping changes (\eg changes a series of colors from red to blue), the visible clues associated with the target series may also change. In this case, the benchmark test will be able to update the questions in a way that maintains the same semantic intent. In our settings, the system can automatically update the question $q$ and answer $y$ to reflect the new mapping, while keeping the underlying analysis tasks and difficulty unchanged.
Therefore, the answer variable $y$ is not regarded as a static label attached to the fixed image; instead, it is answered in relation to the current legend state. More broadly speaking, in the question-answering generation stage, the benchmark test transforms visual instances into diagnostic evaluation units, which link the decomposed design space to specific tasks.
After an intervention, the question templates are re-instantiated from the updated metadata so that visual references and the expected answer follow the current binding, while the semantic target and analytical intent remain fixed.

\subsection{Counterfactual Variant Generation}

A single base unit only shows that a model succeeds or fails on one particular configuration.
To understand whether performance depends on a specific design choice, and whether a correct answer genuinely relies on the legend, we need to compare matched instances that differ in exactly one controlled factor while keeping the difficulty fixed.
Our pipeline supports two complementary types of variants: \emph{configuration-level interventions}, which reveal how design choices such as encoding channel or palette affect performance, and \emph{instance-level interventions}, which test whether a model's answer causally depends on the legend by modifying only the legend of a fixed chart.

\paragraph{Configuration-level interventions.}
Given the same underlying data, the parametrized generation pipeline can produce different base units by changing a single design dimension, \eg, the encoding channel, the color palette, or the chart family, while keeping all other dimensions fixed.
These controlled interventions explore the parameterized design space defined in \Cref{sec:attributes} and form the basis of the controlled diagnostic experiments reported in \Cref{sec:controlled}.

\paragraph{Instance-level interventions.}
Unlike configuration-level interventions, instance-level interventions do not change the global design configuration.
Instead, they start from an instantiated base unit
$b=(s,l,q,y)$ and apply local modifications to an already-instantiated chart while holding the chart content and question intent fixed, such as shuffling the legend entries, moving the legend to an atypical position, or reducing the legend font size.
We organize these interventions into three diagnostic families (\Cref{tab:variant_families}).
Mapping-preserving variants change only the legend presentation while keeping the label--mark mapping fixed.
A reliable reader should therefore keep the same correct answer (Consistency).
Mapping-altering variants change the label--mark binding itself.
The answer should then update with the new mapping (Responsiveness).
Answerability variants hide some or all of the legend, so the model should answer only when sufficient evidence remains and otherwise abstain.

\begin{table*}[t]
\centering
\footnotesize
\setlength{\tabcolsep}{4pt}
\renewcommand{\arraystretch}{1.0}
\setlength{\extrarowheight}{4pt}
\caption{\textbf{Controlled variant families in \data.}
Each family specifies the intervention and the expected model behavior
under the resulting guide condition.}
\begin{tabular}{
    >{\columncolor{morandi2!30}\raggedright\arraybackslash}m{2.5cm}
    >{\columncolor{morandi4!30}\raggedright\arraybackslash}m{3.7cm}
    >{\columncolor{morandi5!30}\raggedright\arraybackslash}m{5.5cm}
    >{\columncolor{morandi3!30}\raggedright\arraybackslash}m{5.2cm}
}
\toprule
\rowcolor{morandi1!42}
\textbf{Variant family}
& \textbf{Operators}
& \textbf{Controlled intervention}
& \textbf{Expected behavior} \\
\midrule

Mapping-preserving & Reorder, Reposition, ScaleDown, Highlight, AddLabels
& Change presentation or legend entries while preserving
the data and label--mark mapping
& Preserve the correct answer and remain stable across Parsing,
Grounding, and Reasoning \\

Mapping-altering & Swap, SimilarColor
& Change the label--mark binding while preserving the
underlying data and semantic target
& Follow the current binding and update the answer when required \\

Answerability
& Mask
& Hide the legend while preserving the data and
question intent
& Answer when sufficient evidence remains; otherwise abstain \\

\bottomrule
\end{tabular}
\label{tab:variant_families}
\end{table*}

\subsection{Implementation Details}
To generate the dataset, users only need to provide a configuration file that specifies the possible values for each dimension in our parametric space, such as chart families, encoding channels, \revise{legend forms,} and question types.
However, exhaustive enumeration of the full combinatorial design space is impractical.
We therefore release a pre-built snapshot of \revise{26689} instances, and each instance reflects base images $\times$ variants $\times$ question-answer pairs derived from \revise{900} base images, where each base image generates multiple counterfactual variants and sampled question-answer pairs.
The released snapshot samples a subset of applicable variants and question-answer pairs for each base chart which covers representative configurations across this space rather than enumerating the full Cartesian product. 
This policy provides broad coverage of the parameterized space at a tractable size. Controlled diagnostic experiments use separate configurations that fix all non-probed dimensions and enumerate the selected factor within a restricted design subspace.
\revise{We screened constructed samples and obtained near-perfect accuracy, which we treat as a validity check that the generated items are well-formed and answerable from the intended mapping.}
For targeted studies, users can modify the configuration file to fix all dimensions except the one under study, and the pipeline automatically produces matched comparison sets.
The generation pipeline and the pre-built snapshot are available at \url{https://github.com/legendbench/legendbench}.

%% file: 4-Experiments.tex
\section{Experiments and Results}
\label{sec:experiments}
This section demonstrates the \data diagnostic protocol on current VLMs.
We first specify the metrics that constitute a capability profile.
We then report that profile to localize where legend understanding fails.
\revise{Finally, we test whether the diagnosed bottlenecks can guide targeted fine-tuning.}
We include both \textbf{general-purpose models} and \textbf{chart-specific models} to test whether chart specialization changes which legend capabilities remain weak.
The general-purpose models include \texttt{LLaVA-HR}, \texttt{Qwen3-32B}, \texttt{Qwen3-8B}, \texttt{Qwen3.5-27B}, \texttt{GLM-4.6V}, \texttt{Gemini-2.5-Flash}, \texttt{GPT-4.1-mini}, and \texttt{GPT-5.6-sol}.
The chart-specific models include \texttt{ChartInstruct}~\cite{masry-etal-2024-chartinstruct}, \texttt{ChartLlama}~\cite{han2023chartllama}, \texttt{TinyChart}~\cite{zhang-etal-2024-tinychart}, and \texttt{ChartMLLM}~\cite{chartmllm}.
All models are evaluated with the same prompt template and hyperparameters.
Full details and per-subfamily scores are provided in the supplementary material.

\begin{table*}[t]
\centering
\small
\setlength{\tabcolsep}{3.4pt}
\renewcommand{\arraystretch}{1.06}
\caption{Overall diagnostic capability profiles of different models on \data. }
\label{tab:overall_capability_profile}
\resizebox{\textwidth}{!}{%
\begin{tabular}{lccccccccccccc}
\toprule
\multirow{2}{*}{Method}
& \multicolumn{2}{c}{Overall}
& \multicolumn{1}{c}{Parsing}
& \multicolumn{2}{c}{Grounding}
& \multicolumn{2}{c}{Reasoning}
& \multicolumn{3}{c}{Counterfactual}
& \multicolumn{3}{c}{Abstention} \\
\cmidrule(lr){2-3}
\cmidrule(lr){4-4}
\cmidrule(lr){5-6}
\cmidrule(lr){7-8}
\cmidrule(lr){9-11}
\cmidrule(lr){12-14}
& QA Acc.
& Chain Acc.
& Guide Read.
& Bind.
& VE
& Reason.
& Reason.$|$Sup.
& Cons.
& Resp.
& CR
& AP
& AR
& AF1 \\
\midrule

GPT-5.6-sol
& \textbf{83.81} & \textbf{52.43} & \textbf{91.12}
& \textbf{84.39} & \textbf{78.31}
& \textbf{74.16} & \underline{84.85}
& \textbf{76.94} & \textbf{80.22} & \textbf{78.55}
& \textbf{99.69} & \textbf{95.33} & \textbf{97.46} \\

Qwen3.5-27B
& \underline{71.62} & 29.15 & \underline{86.19}
& 69.03 & \underline{62.28}
& \underline{56.79} & 74.97
& \underline{64.77} & 57.80 & \underline{61.09}
& \underline{93.42} & 77.34 & \underline{84.62} \\

Gemini-2.5-Flash
& 67.82 & \underline{30.96} & 79.81
& \underline{75.16} & 47.63
& 53.71 & 79.52
& 55.76 & \underline{58.77} & 57.23
& 62.51 & \underline{95.23} & 75.48 \\

Qwen3-32B
& 66.09 & 25.56 & 80.58
& 67.85 & 49.37
& 52.37 & 76.61
& 57.21 & 56.38 & 56.79
& 66.78 & 86.03 & 75.19 \\

GLM-4.6V
& 64.59 & 21.22 & 78.74
& 62.55 & 51.86
& 53.41 & 77.98
& 59.57 & 41.96 & 49.24
& 82.01 & 46.59 & 59.42 \\

GPT-4.1-mini
& 64.19 & 21.25 & 76.82
& 61.46 & 57.69
& 50.38 & 73.02
& 61.15 & 30.95 & 41.10
& 77.98 & 4.14 & 7.86 \\

Qwen3-8B
& 58.02 & 14.17 & 75.53
& 52.90 & 43.57
& 47.09 & 77.61
& 47.92 & 45.98 & 46.93
& 52.40 & 85.74 & 65.04 \\

LLaVA-HR
& 5.78 & 0.07 & 9.78
& 4.82 & 0.29
& 5.33 & 60.00
& 5.14 & 1.41 & 2.21
& 0.00 & 0.00 & 0.00 \\

\midrule

ChartInstruct
& 15.19 & 0.12 & 15.46
& 17.02 & 6.81
& 20.86 & 19.23
& 14.48 & 1.77 & 3.15
& 0.00 & 0.00 & 0.00 \\

TinyChart
& 13.77 & 0.02 & 16.53
& 6.32 & 18.30
& 15.53 & 50.00
& 10.05 & 2.22 & 3.64
& 0.00 & 0.00 & 0.00 \\

ChartLlama
& 12.26 & 0.10 & 16.57
& 4.07 & 6.98
& 22.92 & \textbf{100.00}
& 11.38 & 1.34 & 2.39
& 0.00 & 0.00 & 0.00 \\

ChartMLLM
& 9.07 & 0.10 & 17.20
& 5.00 & 1.30
& 8.10 & 62.50
& 80.70 & 15.80 & 26.43
& 0.00 & 0.00 & 0.00 \\
  
\bottomrule
\end{tabular}%
}
\vspace{2pt}
\parbox{\linewidth}{\scriptsize
\textit{Note.} QA Acc. = QA Accuracy; \revise{Chain Acc. = Chain Accuracy}; Guide Read. = Guide Reading; \revise{Bind. = Binding}; VE = Value Extraction; Reason. = Reasoning; \revise{Reason.\textbar Sup. = Conditional reasoning}; Cons. = Consistency; Resp. = Responsiveness; \revise{CR = Harmonic mean of Consistency--Responsiveness; AP = Abstention Precision; AR = Abstention Recall;} AF1 = Abstention F1. For detailed definitions please refer to \Cref{sec:metric}.
}
\end{table*}

\subsection{Diagnostic Metrics}
\label{sec:metric}
Instead of a single benchmark score, \data reports a capability profile along four views: overall performance, stage-wise performance, counterfactual behavior, and abstention.
Each metric is computed only over the questions to which the metric is applicable. %

\paragraph{Overall performance.}
Overall accuracy reports how often the model is correct before the profile localizes which capability failed.
\textbf{QA Accuracy} is the conventional score, \ie, the proportion of individual questions answered correctly.
\revise{In \data, these individual questions are also assembled into diagnostic chains, each grouping the upstream probes required by a target question.}
\revise{\textbf{Chain Accuracy} is the proportion of chains for which every applicable question is answered correctly.
For example, if the target question is ``What is the value of `Finance' in 2015?'', Chain Accuracy credits the instance only when the model correctly parses the Finance entry, binds it to the corresponding marks, and extracts the value.}

\paragraph{Stage-wise performance.}
These metrics localize where the evidence chain breaks, following the parsing--grounding--reasoning order of T1--T3.
Parsing is scored by \textbf{guide reading} (\texttt{Guide Read.}), which measures whether the model can extract the mapping that the legend makes explicit.
\revise{Grounding comprises two metrics: \textbf{binding} (\texttt{Bind.}) measures whether that mapping can be associated with the corresponding chart mark, and \textbf{value extraction} (\texttt{VE}) measures whether the model recovers the quantitative value of the relevant mark, cell, or scale.}
Reasoning is scored by \textbf{reasoning} (\texttt{Reason.}), which is raw accuracy on questions that require inference.
\revise{To separate inherited evidence failures from genuine reasoning errors, we additionally report \textbf{conditional reasoning} (\texttt{Reason.$|$Sup.}), which restricts reasoning evaluation to cases whose upstream chain is already correct.}

\revise{\paragraph{Counterfactual behavior.}
For each original--variant pair in a counterfactual group, we test whether the prediction tracks the legend as it should.
\textbf{Consistency} (\texttt{Cons.}) checks that the model gives the same correct answer under a mapping-preserving intervention, where
\textbf{Responsiveness} (\texttt{Resp.}) checks that it updates to the new correct answer under a mapping-altering one.
\textbf{CR} is the harmonic mean of the two, and is high only when the model is both stable under mapping-preserving changes and responsive under mapping-altering ones.}

\paragraph{\revise{Abstention.}}
\label{par:abstention-metrics}
\revise{Since most chart questions remain answerable, scoring the abstain/answer decision with plain accuracy barely penalizes a model that never refuses.
We therefore focus on whether the model abstains when it cannot answer.}
\revise{\textbf{Abstention Precision} (\texttt{AP}) measures the proportion of the model's refusal decisions that are correct, i.e., the fraction of refusals made on genuinely unanswerable questions.
\textbf{Abstention Recall} (\texttt{AR}) measures the proportion of genuinely unanswerable questions on which the model correctly refuses.
\textbf{Abstention F1} (\texttt{AF1}) is the harmonic mean of Abstention Precision and Abstention Recall, balancing over-refusal and under-refusal.}

\revise{Together, these four complementary views characterize the models’ capabilities and form a comprehensive capability profile.}

\subsection{Capability Profile}
Table~\ref{tab:overall_capability_profile} reports the capability profile of models.

\textbf{Overall performance.}
\revise{It is evident that QA Accuracy is substantially higher than Chain Accuracy for every model.
This gap indicates that a correct final answer does not necessarily reflect successful completion of the underlying legend-understanding process. 
A model may arrive at the correct answer despite failures in legend parsing, legend-mark binding, value extraction, or downstream reasoning. 
Thus, conventional QA accuracy can mask substantial failures in intermediate capabilities. 
We also find that even on these standardized charts, the strongest evaluated model, \texttt{GPT-5.6-sol}, reaches only $52.43\%$ Chain Accuracy, \textbf{leaving substantial room for improvement}.}

\textbf{Stage-wise performance.}
We next localize failures along the diagnostic stages in the capability profile.
Parsing is the strongest stage for the stronger general-purpose models, which typically extract categorical labels, direct labels, and continuous scale elements that the legend presents explicitly.
\revise{\texttt{GPT-5.6-sol} reaches $91.12\%$ Guide Reading, and \texttt{Gemini-2.5-Flash}, \texttt{Qwen3-32B}, \texttt{GPT-4.1-mini}, and \texttt{Qwen3.5-27B} remain between $75\%$ and $86\%$.}
Basic legend understanding is therefore not their main capability boundary.
\revise{In contrast, the chart-specific models remain near $16\%$ at this stage, and \texttt{LLaVA-HR} shows the same early breakdown, so those failures begin with acquiring legend evidence rather than only at later stages.}
Performance then generally drops from parsing to grounding.
\revise{Even \texttt{GPT-5.6-sol} falls to $84.39\%$ Binding and $78.31\%$ Value Extraction.}

\revise{The failed location of that drop is model-dependent: \texttt{Qwen3.5-27B} already loses much of its accuracy at binding ($86.19\%$ to $69.03\%$), whereas Qwen3-8B shows a sharper grounding bottleneck, falling from 75.53\% Guide reading to 52.90\% Binding and 43.57\% Value Extraction.
By contrast, \texttt{Gemini-2.5-Flash} remains comparatively strong at binding ($75.16\%$) but falls to $47.63\%$ on value extraction.}
Reading the legend therefore does not guarantee that the model can associate labels with marks or recover quantitative values.
The chart-specific models remain weak across both grounding metrics, indicating a broader inability to turn legend information into usable visual and numerical evidence. 
\revise{Raw reasoning remains below the grounding stages for the strong general-purpose models.
However, conditional reasoning is consistently higher than raw reasoning once the upstream chain is correct, for example $84.85\%$ versus $74.16\%$ for \texttt{GPT-5.6-sol} and $79.52\%$ versus $53.71\%$ for \texttt{Gemini-2.5-Flash}.
Many apparent reasoning errors are therefore inherited from earlier evidence failures rather than from the final inference step itself.
Conditional scores for models with near-zero Chain Accuracy, such as \texttt{ChartLlama} at $100.00\%$ Reason.$|$Sup., should not be read as evidence of stable reasoning ability. This score is computed only on the small subset of instances where the model first obtains the correct support, so even a single correct reasoning decision can yield an apparently perfect conditional score.}

\textbf{Counterfactual behavior.}
We next test whether predictions track legend interventions rather than a fixed chart pattern.
\revise{\texttt{GPT-5.6-sol} is comparatively balanced, with $76.94\%$ Consistency, $80.22\%$ Responsiveness, and $78.55\%$ CR.
This balance is not shared by all models with moderate QA Accuracy.
\texttt{GPT-4.1-mini} reaches $61.15\%$ Consistency but only $30.95\%$ Responsiveness, so it often fails to update the answer when the legend-binding changes.}
The chart-specific models remain near floor on Responsiveness and CR, indicating that their predictions are not grounded in the current legend specification.
These failures are invisible to standard accuracy, which evaluates each chart in isolation.

\textbf{Abstention.}
We next examine whether models refuse when the required legend evidence is missing.
\revise{\texttt{GPT-5.6-sol} and \texttt{Gemini-2.5-Flash} are strong on Abstention Precision (AP), Abstention  Recall (AR) and Abstention F1 (AF1). In contrast, earlier models like \texttt{GPT-4.1-mini} reach $64.19\%$ QA Accuracy with only $7.86\%$ AF1, showing that strong aggregate accuracy does not necessarily translate into reliable abstention.}
\revise{All chart-specific models, as well as \texttt{LLaVA-HR}, score $0.00$ on AP, AR and AF1, indicating that they do not abstain correctly under missing evidence.}
Chart-specific fine-tuning therefore does not by itself produce calibrated refusal.
Abstention is thus a distinct failure mode: a model can answer many remaining questions while still failing to recognize when the visible legend information is no longer sufficient.

\subsection{\revise{Diagnosis-Guided Fine-tuning}}
\label{sec:finetuning}

\paragraph{\revise{Motivation.}}
\revise{The capability profile indicates where the failure occurs.
This raises a practical question: can such fine-grained diagnosis guide targeted fixes, rather than blanket retraining?
}

\paragraph{\revise{Experimental design.}}
\revise{
We conduct a small-scale fine-tuning study using \texttt{Qwen3-8B} as the base model.
We use \texttt{Qwen3-8B} rather than a larger backbone because resource-constrained settings, such as on-device deployment, make targeted repair of a small open-weight model more relevant than further scaling.
We compare two fine-tuning settings against the base model.
\textit{Random-LB FT} randomly samples instances and tests whether simply adding benchmark-style data is sufficient.
\textit{Bottleneck-LB FT} uses the capability profile of the base model to select training instances from its weakest capability regions. %
All the fine-tuning settings use the same model, training budget, and evaluation set, and are evaluated on the held-out \data test set.
We additionally evaluate on ChartQA~\cite{chartqa} to test whether fine-tuning gains transfer beyond \data, rather than remaining specific to held-out items from the same generator.} %

\paragraph{\revise{Results.}
}
\revise{
\Cref{tab:diagnostic_finetuning} shows that diagnosis-guided fine-tuning yields larger gains than random sampling, achieving the highest QA Accuracy and Chain Accuracy.
The contrast is sharper on transfer, where \textit{Random-LB FT} improves held-out \data but degrades ChartQA.
In contrast, \textit{Bottleneck-LB FT} raises ChartQA to 
84.32\%, indicating that targeting the diagnosed bottlenecks improves the underlying capability rather than fitting to \data's distribution.}

\begin{table}[t]
\centering
\small
\setlength{\tabcolsep}{4.0pt}
\renewcommand{\arraystretch}{1.1}
\caption{Fine-tuning results on \data. Bottleneck-guided and random \data fine-tuning improve the base model. The bottleneck-guided setting uses a strict allocation ratio emphasizing identity binding and value extraction.}
\label{tab:diagnostic_finetuning}
\begin{tabular}{@{}lccccccc@{}}
\toprule
Setting
& ChartQA
& QA Acc.
& Chain Acc.
& Guide
& Bind.
& VE
& Reason. \\
\midrule
Base
& 83.80 & 58.00 & 14.17 & 75.53 & 52.90 & 43.57 & 47.09 \\
Random
& 82.65 & 64.52 & 21.70 & 80.30 & 61.50 & 49.00 & \textbf{55.10} \\
Bottleneck
& \textbf{84.32} & \textbf{67.68} & \textbf{24.60}
& \textbf{83.60} & \textbf{69.00} & \textbf{50.70} & 52.10 \\
\bottomrule
\end{tabular}
\end{table}

%% file: 4-2-Controlled.tex
\section{Controlled Diagnostic Experiments}
\label{sec:controlled}
\revise{Section~\ref{sec:experiments} evaluates models on a pre-built snapshot of the parameterized space rather than on a strictly controlled factorial design.
This is because the full combinatorial space is too large to be enumerated, and the resulting capability profiles should be read as preliminary patterns.
Nevertheless, these profiles can still provide informative findings that serve as a starting point.
We then use the generation pipeline of \data to generate matched counterfactual groups for strictly controlled diagnostic experiments, varying one factor while holding the chart data, question intent, and remaining visual settings fixed.
An online interactive demo\footnote{\url{https://github.com/legendbench/legendbench}} lets readers explore the snapshot, identify additional patterns, and generate configurations for the same controlled follow-up.
We report three such studies: encoding-channel effects, legend-order shortcuts, and abstention under varying visibility.
Unless otherwise noted, the main text reports \textsc{Qwen3.5-27B}, with results for the other evaluated models in the supplementary material.}

\subsection{\revise{Are Failures Invariant to Encoding Channel?}} %

\paragraph{Motivation.}
\revise{The capability profile in Section~\ref{sec:experiments} shows a persistent gap between legend parsing and legend grounding.
That aggregate gap, however, mixes encoding channel with other visual factors.
This study asks whether different channels affect the gap, and whether a different encoding choice, including redundant double-channel encodings, can close it.
We therefore isolate encoding channel while holding the chart data, question intent, and remaining visual settings fixed.}

\paragraph{Experimental design.}
We organize this study by \textbf{mark type} rather than by chart family because channel effectiveness is fundamentally tied to the underlying mark.
Different charts can share the same mark type and therefore use the same channels to distinguish series.
For point marks (scatter plots), the series identity can be encoded by \textit{color} or \textit{marker}.
For line marks (line charts and radar charts), the series identity can be encoded by \textit{color}, \textit{marker}, or \textit{line style}.
For area marks (bar, area, and pie charts), the series identity can be encoded by \textit{color} or \textit{texture}.
In addition to testing single-channel encoding, we also test double-channel encoding to investigate whether redundant cues can compensate for weak single-channel conditions.

\paragraph{Results on point marks.}
\Cref{tab:channel_by_mark} (a) shows that \textit{marker} achieves the highest parsing accuracy ($\approx 0.9$), followed by \textit{color} ($\approx 0.8$), while \textit{color+marker} ($\approx 0.85$) provides no apparent additional gain over marker alone.
This suggests that discrete point marks offer locally distinctive cues that models can match to legend entries without long-range tracking, making marker alone a sufficiently strong channel.
\begin{table}[h]
    \fontsize{8.5}{11}\selectfont
    \centering
    \caption{Parsing, grounding, and reasoning accuracy of \textsc{Qwen3.5-27B} when only the series-identity channel varies. Cell shading indicates performance (darker = higher).}
    \label{tab:channel_by_mark}

    \textbf{(a) Point Marks (Scatter)   }\\[0.3em]
    \begin{minipage}{0.87\columnwidth}%
        \centering
        \renewcommand{\arraystretch}{1.15}
        \setlength{\arrayrulewidth}{0.5pt}
        \begin{tabularx}{\linewidth}{lXXX}
            \hline
            \rowH\textbf{Ch.} & \centering\small\textbf{Parsing} & \centering\small\textbf{Grounding} & \centering\arraybackslash\small\textbf{Reasoning} \\ \hline
            \rowH Color           & \centering\acc{0.80} & \centering\acc{0.83} & \centering\arraybackslash\acc{0.93} \\ \hline
            \rowH Marker          & \centering\acc{0.88} & \centering\acc{0.83} & \centering\arraybackslash\acc{1.00} \\ \hline
            \rowH Color+Marker    & \centering\acc{0.85} & \centering\acc{0.92} & \centering\arraybackslash\acc{1.00} \\ \hline
        \end{tabularx}
    \end{minipage}%
    \begin{minipage}{0.13\columnwidth}\colorbar\end{minipage}
    \label{tab:PointMarks}

    \textbf{(b) Line Marks (Line/Radar)   }\\[0.3em]
    \begin{minipage}{0.87\columnwidth}%
        \centering
        \renewcommand{\arraystretch}{1.15}
        \setlength{\arrayrulewidth}{0.5pt}
        \begin{tabularx}{\linewidth}{lXXX}
            \hline
            \rowH\textbf{Ch.} & \centering\small\textbf{Parsing} & \centering\small\textbf{Grounding} & \centering\small\arraybackslash\textbf{Reasoning} \\ \hline
            \rowH Color  & \centering\acc{0.84} & \centering\acc{0.86} & \centering\arraybackslash\acc{0.89} \\ \hline
            \rowH Marker  & \centering\acc{0.90} & \centering\acc{0.80} & \centering\arraybackslash\acc{0.90} \\ \hline
            \rowH Line Style  & \centering\acc{0.83} & \centering\acc{0.76} & \centering\arraybackslash\acc{0.83} \\ \hline
            \rowH Color+Marker & \centering\acc{1.00} & \centering\acc{0.83} & \centering\arraybackslash\acc{0.92} \\ \hline
            \rowH Color+Line Style & \centering\acc{0.93} & \centering\acc{0.82} & \centering\arraybackslash\acc{0.89} \\ \hline
            \rowH Line Style+Marker & \centering\acc{0.94} & \centering\acc{0.80} & \centering\arraybackslash\acc{0.88} \\ \hline
        \end{tabularx}
    \end{minipage}%
    \begin{minipage}{0.13\columnwidth}\colorbar\end{minipage}
    \label{tab:LineMarks}

    \textbf{(c) Area Marks (Bar/Pie/Area)   }\\[0.3em]
    \begin{minipage}{0.87\columnwidth}%
        \centering
        \renewcommand{\arraystretch}{1.15}
        \setlength{\arrayrulewidth}{0.5pt}
        \begin{tabularx}{\linewidth}{lXXX}
            \hline
            \rowH\textbf{Ch.} & \centering\small\textbf{Parsing} & \centering\small\textbf{Grounding} & \centering\arraybackslash\small\textbf{Reasoning} \\ \hline
            \rowH Color  & \centering\acc{1.00} & \centering\acc{0.80} & \centering\arraybackslash\acc{0.87} \\ \hline
            \rowH Texture  & \centering\acc{0.83} & \centering\acc{0.50} & \centering\arraybackslash\acc{0.70} \\ \hline
            \rowH Color+Texture & \centering\acc{0.96} & \centering\acc{0.72} & \centering\arraybackslash\acc{0.87} \\ \hline
        \end{tabularx}
    \end{minipage}%
    \begin{minipage}{0.13\columnwidth}\colorbar\end{minipage}
    \label{tab:AreaMarks}
\end{table}

\paragraph{Results on line marks.}
\revise{As shown in \Cref{tab:channel_by_mark}(b), single-channel accuracy is not uniform: \textit{color} remains comparatively strong (grounding accuracy $>0.86$), while \textit{line style} is weaker ($\approx 0.8$).}
This indicates that for line marks, marker and line style differences can be difficult to distinguish because they are rendered at a smaller scale.
Notably, the double encoding \textit{color+marker} achieves the highest accuracy overall, suggesting that combining a globally salient channel (color) with a locally distinctive one (marker) provides complementary cues that together yield more robust legend grounding.

\paragraph{Results on area marks.}
As shown in \Cref{tab:channel_by_mark}(c), models remain strong under \textit{color}, with all  three accuracies above 0.8, but degrade sharply under \textit{texture}, where grounding accuracy drops to 0.50.
Unlike color, which provides a single globally distinctive cue, textures such as hatching, dots, and crosshatch require the model to discriminate fine-grained structural details. \revise{The largest performance loss occurs at the Grounding stage, indicating that texture identity is not preserved between the small legend and the corresponding area mark.}
Compared with color-only encoding, \textit{color+texture} leads to a noticeable performance drop, indicating that the extra texture channel appears to introduce additional perceptual complexity, making legend grounding more difficult.

\paragraph{Takeaway.}

\revise{Overall, legend understanding is not invariant to encoding channel.
Color yields the most effective and reliable grounding performance.
Marker and line style can complement color for point and line marks, whereas texture introduces substantial perceptual difficulty and exposes the transition from successful parsing to failed grounding.
Thus, encoding channel serves as a diagnostic factor for identifying stage-specific bottlenecks in model performance rather than a universally beneficial design choice.}

\subsection{\revise{Do Models Rely on Legend Order as a Shortcut?}}
\paragraph{\revise{Motivation.}}
\revise{
Legend entries are sometimes ordered to mirror the visual ordering of series at a salient location, such as the final $y$-position, making cross-referencing easier.
However, this alignment also creates a potential shortcut: instead of locating the target series and determining its rank in the chart, a model may simply return the target label's position in the legend. 
We therefore test whether model performance depends on this alignment and whether answers follow the target's legend-entry position once it is decoupled from visual rank.
}

\paragraph{\revise{Experimental design.}}
\revise{
We manipulate the overall legend order to vary each target's legend-entry position while holding all other factors fixed.
For each base chart with $N$ labels, we generate $N$ cyclic reorderings so that every target label appears once at each legend-entry position from $1$ to $N$. 
Across variants, the underlying data, chart geometry, label--mark mappings, physical legend location, target labels, and questions remain unchanged. 
The expected answer is determined independently of legend order from the chart data or geometry using chart-type-specific ordering rules, and therefore remains fixed across all reorderings.
This construction yields aligned conditions, in which the target's legend-entry position matches its visual rank, and conflict conditions, where the two differ.
}

\paragraph{\revise{Results.}}
\revise{We first compare accuracy on aligned charts, where legend-entry position matches the gold visual rank, with conflict charts, where the two differ.
Accuracy drops from 0.58 to 0.30, indicating reliance on this legend-order shortcut.
We further analyze conflict charts and find that 0.30 of predictions recover the gold visual rank, while 0.36 follow the current legend-entry position, indicating a systematic pull toward legend-entry position.
This pull becomes more explicit across rotations, where 16\% of targets follow the legend order in every rotation despite an unchanged gold rank.
These results indicate that the performance drop under conflict is partly driven by legend-entry position substitution rather than solely by nonspecific grounding errors.
}

\paragraph{\revise{Takeaway.}} 
\revise{These results reveal a legend order bias: when legend-entry position and visual rank conflict, \textsc{Qwen3.5-27B} often substitutes the former for the latter.
From a chart-design standpoint, if no competing constraint requires otherwise, aligning legend order with visual rank is a reasonable default, because this convention eases human cross-referencing and also improves model accuracy in our setting.
For models, however, such alignment should not replace the underlying capability.
Robust legend understanding still requires visual-rank grounding rather than reliance on this shortcut.
}

\subsection{\revise{How Does Visibility Affect Abstention?}}
\paragraph{Motivation.}

\revise{The capability profile shows that current VLMs perform strongly on aggregate legend-aware abstention. 
This finding motivates a more specific question: can this aggregate strength withstand fine-grained scrutiny under partial legend visibility, where models must distinguish evidence that is sufficient for the current query from evidence that is not? }
We therefore examine how legend visibility affects answering behavior, and under which evidence-availability conditions current VLMs answer rather than abstain.

\paragraph{Experimental design.}
We isolate legend visibility as the independent variable by implementing four occlusion policies: \texttt{Mask\_Full} (the entire legend region is removed), \texttt{Keep\_Target} (only the target entry is shown), \texttt{Hide\_Target} (only the target entry is removed), and \texttt{Keep\_Distractor} (only a randomly sampled non-target entry is shown).
For each condition, the expected answer is retained if the remaining visible entries are sufficient to resolve the query, and set to ``I DON'T KNOW'' otherwise.

\paragraph{Results.}

\revise{\Cref{tab:MaskOcclusion_metric} reports \textbf{Abstention Precision (AP)}, \textbf{Abstention Recall (AR)}, and \textbf{Abstention F1 (AF1)} under different legend-visibility conditions.}
A clear contrast emerges between \emph{complete absence of legends} and \emph{partial observability}.
\revise{Under \texttt{Mask\_Full}, the model achieves high \texttt{AP} ($0.94$) but low \texttt{AR} ($0.18$), yielding an \texttt{AF1} of $0.30$.}
A similar pattern appears under \texttt{Keep\_Distractor}, where only irrelevant legend entries remain visible: \texttt{AP} remains high, while \texttt{AR} falls to $0.11$ and \texttt{AF1} to $0.20$.
These results indicate that when the required legend evidence is entirely absent or obviously irrelevant, the model can sometimes detect the lack of usable information.
In contrast, partial visibility creates a much harder rejection problem. 
Under \texttt{Keep\_Target}, the model fails to recognize unanswerable questions, as reflected by its low \texttt{AF1} of $0.04$.
\revise{The most difficult condition is \texttt{Hide\_Target}, where the model almost never abstains when it should, resulting in an \texttt{AF1} of only $0.03$.}
Once the target mapping is removed but other legend entries remain visible, the model rarely identifies the need to abstain. Instead, it tends to continue guessing under misleading partial evidence.

\paragraph{Takeaway.}
Legend-aware abstention depends strongly on evidence availability.
\revise{Under complete removal or irrelevant-only visibility, refusals are usually correct when they occur, but recall remains low. 
Partial visibility is more problematic: \texttt{Keep\_Target} and \texttt{Hide\_Target} produce near-zero abstention recall and F1.}
This shows that the main difficulty is not detecting obvious absence, but judging whether partial evidence is still sufficient for answering.

\begin{table}[h]
    \centering
    \caption{Abstention of \textsc{Qwen3.5-27B} when only legend visibility varies.
    } %

    \begin{minipage}{0.87\columnwidth}%
        \centering
        \renewcommand{\arraystretch}{1.15}
        \setlength{\arrayrulewidth}{0.5pt}
        \begin{tabularx}{\linewidth}{lXXX}
            \hline
            \rowH\textbf{Condition} & \centering\textbf{AP} & \centering\textbf{AR} & \centering\arraybackslash\textbf{AF1} \\ \hline
            \rowH Mask\_Full         & \centering\acc{0.94} & \centering\acc{0.18} & \centering\arraybackslash\acc{0.30} \\ \hline
            \rowH Keep\_Target       & \centering\acc{0.33} & \centering\acc{0.02} & \centering\arraybackslash\acc{0.04} \\ \hline
            \rowH Hide\_Target       & \centering\acc{0.05} & \centering\acc{0.02} & \centering\arraybackslash\acc{0.03} \\ \hline
            \rowH Keep\_Distractor   & \centering\acc{0.98} & \centering\acc{0.11} & \centering\arraybackslash\acc{0.20} \\ \hline
        \end{tabularx}
    \end{minipage}%
    \begin{minipage}{0.13\columnwidth}\colorbar\end{minipage}
    \label{tab:MaskOcclusion_metric}
\end{table}

%% file: 5-Discussion.tex
\section{Discussion}

\paragraph{Move Beyond Static Accuracy.}
A broader takeaway is that static benchmark accuracy alone is often insufficient for evaluating genuine chart understanding.
By combining a parameterized space with matched counterfactual variants, \data enables a more diagnostic evaluation of whether model predictions are causally grounded in legend information rather than supported by shortcut cues.
This design naturally supports a two-stage evaluation workflow, where users can first run the evaluation to get a \revise{capability profile,} and then conduct controlled diagnostic experiments to \revise{isolate the factor that caused the failure.}
We believe that this evaluation workflow may extend beyond legends to other chart components, and even to other visual reasoning tasks that are similarly vulnerable to shortcut exploitation.

\revise{\paragraph{Guiding Principles.}
We distill four practical principles for diagnosis, model selection, legend-condition testing, and targeted repair.
First, localize the failed stage before collecting more data.
Parsing, grounding, reasoning, and abstention fail for different reasons, and aggregate accuracy hides that distinction.
A useful starting point is a capability profile that separates isolated QA success from complete-chain success, and mapping-preserving consistency from mapping-altering responsiveness.
Second, select models by the capability the intended workflow actually requires, not by overall QA accuracy.
A model that answers many questions correctly can still fail at legend-mark binding, mapping-altering updates, or abstention under missing evidence.
When the intended workflow must limit answering without sufficient evidence, it can be appropriate to trade QA accuracy for higher abstention recall.
Third, while human interpretability remains the primary design objective, a chart can additionally be made more model-reliable, provided that the choice does not conflict with human-centered design.
For example, our study showed that color is the most reliable grounding channel for current VLMs, and aligned legend order already eases human cross-referencing while also improving model accuracy.
Even so, improving the model's genuine legend understanding remains the primary goal, and more favorable encodings and legend-order conventions should not replace visual-rank grounding.
Fourth, when repairing a diagnosed failure, train on bottleneck samples rather than on a uniform draw from the benchmark.
Bottleneck-guided fine-tuning improves the localized weaknesses more than random sampling from the same generator, and the resulting gains transfer beyond \data.
}

\paragraph{Hallucination under Partial Legend Evidence.}
Our abstention experiments reveal that many models, especially chart-specific models, frequently hallucinate answers when the required legend evidence is removed.
The controlled experiment further suggests that this phenomenon is more severe when the evidence is only partially available rather than entirely absent.
In the context of chart understanding, the costs of the two error types are asymmetric: failing to answer a solvable question is merely unhelpful, while confidently producing an unfounded answer can mislead downstream decisions that rely on the reported data.
Future progress should therefore focus on models that can more reliably assess whether the available evidence is sufficient, or at least refuse to answer when they are uncertain.

\paragraph{Diversity and Control.}
We improve visual diversity of \data by combining multiple data sources (adapted ChartBench and synthetic generation) and by varying factors such as chart themes and backgrounds.
Nevertheless, template-based controlled generation inevitably introduces a degree of visual homogeneity.
A promising future direction is to generate counterfactual variants from in-the-wild visualizations, which would greatly improve diversity.
However, this would likely require recovering the underlying data and legend specification from rendered charts, which is a more challenging problem.
Moreover, the extent to which such recovered representations can support precise and semantically consistent interventions remains unclear.
Balancing visual diversity with the level of controllability required for causal diagnosis is therefore an important challenge for future work.

\revise{\paragraph{Public Benchmark Contamination.}
We have already released both the snapshot and the generation pipeline.
It is unavoidable that the leakage of posted items can inflate accuracy on those exact instances.
Although new matched groups can still be generated from different seeds and configurations, regeneration cannot guarantee independence from the public materials.
Public-generator contamination therefore remains a limitation for long-term use of frozen snapshot scores.
However, we would like to clarify that the reusable contribution of \data is a diagnostic instrument for localizing capability failures, rather than a protected exam.
The capability-task taxonomy identifies where legend understanding breaks, and matched interventions test whether an answer tracks the declared mapping rather than a shortcut.
The relevant signal is therefore the capability profile, not a single snapshot ranking.
The diagnosis-guided fine-tuning study supports this use of the instrument.
Training on diagnosed bottleneck samples repairs the localized failures more effectively than random sampling from the same generator, and the resulting gains transfer beyond \data to ChartQA (\Cref{tab:diagnostic_finetuning}).
These results suggest that the diagnostic use of \data remains useful for developing more reliable chart understanding models by exposing capability gaps and producing targeted training data.
}

%% file: 6-Conclusion.tex
\section{Conclusion}
\label{sec:conclusion}
We introduced \data, a diagnostic benchmark for legend understanding with counterfactual interventions.
Rather than evaluating chart understanding solely through aggregate accuracy, \data organizes evaluation around four legend-centric capability families and a parameterized generation pipeline that enables matched counterfactual testing. 
Our experiments reveal persistent bottlenecks in reliable legend-mark binding, counterfactual consistency, and abstention under incomplete evidence.
\revise{The fine-tuning study shows that the same profile can guide targeted repair: training on diagnosed bottleneck samples improves binding and value extraction more than training on already-strong capabilities.}
Beyond a capability profile, the controlled diagnostic experiments indicate that how factors such as encoding channel, \revise{legend order}, and legend visibility change where the evidence chain breaks.
We hope \data can serve as a useful \revise{diagnostic} testbed for developing more robust and genuinely grounded chart understanding systems.